\documentclass[10pt,twocolumn,letterpaper]{article}

\usepackage{cvpr}
\usepackage{times}
\usepackage{epsfig}
\usepackage{graphicx}
\usepackage{amsmath}
\usepackage{amssymb}

\usepackage{pifont}
\newcommand{\cmark}{\ding{51}}%
\newcommand{\xmark}{\ding{55}}%
\usepackage{multirow}
\newcommand{\RNum}[1]{\uppercase\expandafter{\romannumeral #1\relax}}

\usepackage{booktabs}
\usepackage{tabularx}
\usepackage{multirow}
\usepackage[ruled]{algorithm2e}
\newcommand{\STAB}[1]{\begin{tabular}{@{}c@{}}#1\end{tabular}}

\usepackage{subfig}

\usepackage[acronym,toc]{glossaries} 
\newglossaryentry{VAE}{name=VAE,description={Variational Autoencoder},first={\glsentrydesc{VAE} (\glsentrytext{VAE})}}
\newglossaryentry{AE}{name={AE},description={Autoencoder},first={\glsentrydesc{AE} (\glsentrytext{AE})},plural=AEs,descriptionplural={Autoencoders},firstplural={\glsentrydescplural{AE} (\glsentryplural{AE})}}
\newglossaryentry{AAE}{name={AAE},description={Augmented Autoencoder},first={\glsentrydesc{AAE} (\glsentrytext{AAE})},plural=AAEs,descriptionplural={Augmented Autoencoders},firstplural={\glsentrydescplural{AAE} (\glsentryplural{AAE})}}
\newglossaryentry{CVAE}{name=CVAE,description={Conditional Variational Autoencoder},first={\glsentrydesc{CVAE} (\glsentrytext{CVAE})}}
\newglossaryentry{SSD}{name={SSD},description={Single Shot Multibox Detector},first={\glsentrydesc{SSD} (\glsentrytext{SSD})}}
\newglossaryentry{CNN}{name=CNN,description={Convolutional Neural Network},first={\glsentrydesc{CNN} (\glsentrytext{CNN})}, plural=CNNs,descriptionplural={Convolutional Neural Networks},firstplural={\glsentrydescplural{CNN} (\glsentryplural{CNN})}}
\newglossaryentry{SIFT}{name=SIFT,description={Scaled Invariant Feature Transform},first={Scaled Invariant Feature Transform (SIFT)}}
\newglossaryentry{SURF}{name=SURF,description={Speeded Up Robust Features},first={Speeded Up Robust Features (SURF)}}
\newglossaryentry{PnP}{name=PnP,description={Perspective-n-Point}, first={Perspective-n-Point (PnP)}}
\newglossaryentry{RF}{name=RF,description={Random Forrest},first={Random Forest (RF)},plural=RFs,descriptionplural={Random Forests},
	firstplural={\glsentrydescplural{RF} (\glsentryplural{RF})}}
\newglossaryentry{RANSAC}{name=RANSAC,description={Random Sample Consensus},first={RANSAC},}
\newglossaryentry{PCA}{name=PCA,description={Principal Component Analysis},first={Principal Component Analysis (PCA)}}
\newglossaryentry{LIDAR}{name={LIDAR},description={Light Detection And Ranging},first={\glsentrydesc{LIDAR} (\glsentrytext{LIDAR})}}
\newglossaryentry{kNN}{name={kNN},description={k-Nearest-Neighbor},first={\glsentrydesc{kNN} (\glsentrytext{kNN})}} 
\newglossaryentry{MLP}{name={MLP},description={Multilayer Perceptron},first={\glsentrydesc{MLP} (\glsentrytext{MLP})}, plural=MLPs,descriptionplural={Multilayer Perceptrons},firstplural={\glsentrydescplural{MLP} (\glsentryplural{MLP})}}
\newglossaryentry{EM}{name={EM},description={Expectation Maximization},first={\glsentrydesc{EM} (\glsentrytext{EM})}}
\newglossaryentry{6DOF}{name={6DOF},description={six degrees of freedom},first={\glsentrydesc{6DOF} (\glsentrytext{6DOF})}}
\newglossaryentry{ICP}{name={ICP},description={Iterative Closest Point},first={\glsentrydesc{ICP} (\glsentrytext{ICP})}}
\newglossaryentry{KL}{name={KL},description={Kullback-Leibler},first={\glsentrydesc{KL} (\glsentrytext{KL})}}
\newglossaryentry{VSD}{name={$err_{vsd}$},description={Visible Surface Discrepancy},first={\glsentrydesc{VSD} (\glsentrytext{VSD})}}
\newglossaryentry{mAP}{name={mAP},description={mean Average Precision},first={\glsentrydesc{mAP} (\glsentrytext{mAP})}}
\newglossaryentry{DA}{name={DA},description={Domain Adaptation},first={\glsentrydesc{DA} (\glsentrytext{DA})}}
\newglossaryentry{DR}{name={DR},description={Domain Randomization},first={\glsentrydesc{DR} (\glsentrytext{DR})}}
\newglossaryentry{DL}{name={DL},description={Deep Learning},first={\glsentrydesc{DL} (\glsentrytext{DL})}}
\newglossaryentry{GAN}{name={GAN},description={Generative Adversarial Network},first={\glsentrydesc{GAN}(\glsentrytext{GAN})},plural=GANs,descriptionplural={Generative Adversarial Networks},firstplural={\glsentrydescplural{GAN} (\glsentryplural{GAN})}}
\newglossaryentry{PPF}{name={PPF},description={Point Pair Features},first={\glsentrydesc{PPF} (\glsentrytext{PPF})}}

\usepackage[pagebackref=true,breaklinks=true,letterpaper=true,colorlinks,bookmarks=false]{hyperref}

 \cvprfinalcopy 

\def\cvprPaperID{8404} 

\ifcvprfinal\fi
\begin{document}

\title{Multi-path Learning for Object Pose Estimation Across Domains}

\author{
	Martin Sundermeyer$^{1,2}$, Maximilian Durner$^{1,2}$, En Yen Puang$^{1}$, Zoltan-Csaba Marton$^{1}$, \\ Narunas Vaskevicius$^{3}$, Kai O. Arras$^{3}$,  Rudolph Triebel $^{1,2}$ \vspace{3pt}\\
	$^1$German Aerospace Center (DLR), $^2$Technical University of Munich (TUM), \\ $^3$Robert Bosch GmbH\\
}
%
%
%

\maketitle
\begin{abstract}
	We introduce a scalable approach for object pose estimation trained on simulated RGB views of multiple 3D models together. We learn an encoding of object views that does not only describe an implicit orientation of all objects seen during training, but can also relate views of untrained objects. Our single-encoder-multi-decoder network is trained using a technique we denote "multi-path learning":
	While the encoder is shared by all objects, each decoder only reconstructs views of a single object. Consequently, views of
	different instances do not have to be separated in the latent space and can share common features. The resulting encoder generalizes well from synthetic to real data and across various instances, categories, model types and datasets. We systematically investigate the learned encodings, their generalization, and iterative refinement strategies on the ModelNet40 and T-LESS dataset. 
	Despite training jointly on multiple objects, our 6D Object Detection pipeline achieves state-of-the-art results on T-LESS at much lower runtimes than competing approaches. \footnote{Code can be found here: \url{https://github.com/DLR-RM/AugmentedAutoencoder/tree/multipath}}
\end{abstract}

\section{Introduction}
Object pose estimation, i.e. estimating the 3D rotation and translation, is a cornerstone in many perception related applications like augmented reality or robotics.
For many years this field was dominated by template- and feature-based approaches.
Thereby, a given object model or a set of extracted object features is matched into the underlying sensor data.
Depending on the quality of the data and the difficulty of the scene, these approaches are still quite competitive. \cite{Hodan_2018_ECCV}
In recent years however, machine learning, and in particular~\gls{DL} techniques have emerged as another key approach \cite{}. 
They are often computationally efficient and can show
higher robustness towards sensor noise, clutter and environment changes \cite{zakharov2019dpod, Sundermeyer2019, wang2019densefusion}.
On the other hand, the need for pose-annotated data as well as lengthy training phases make them less flexible compared to traditional methods. To tackle the lack of annotations, several methods have recently shifted to train on synthetic data rendered from 3D models \cite{kehl2017ssd, sundermeyer2018implicit, tremblay2018deep}. However, it is still common to train individual models for every newly encountered instance which is not scalable. Attempts to train on many objects at once often result in deteriorated performance \cite{zakharov2019dpod}. Furthermore, due to object discriminative training, most of these approaches are not suited to generalize to untrained objects.
Since the real world consists of large amounts of object categories and instances, we propose a more adaptive and scalable approach in this paper.

\begin{figure} 
	\centering
	\captionsetup{width=1.0\columnwidth}
	\includegraphics[width=1.0\columnwidth]{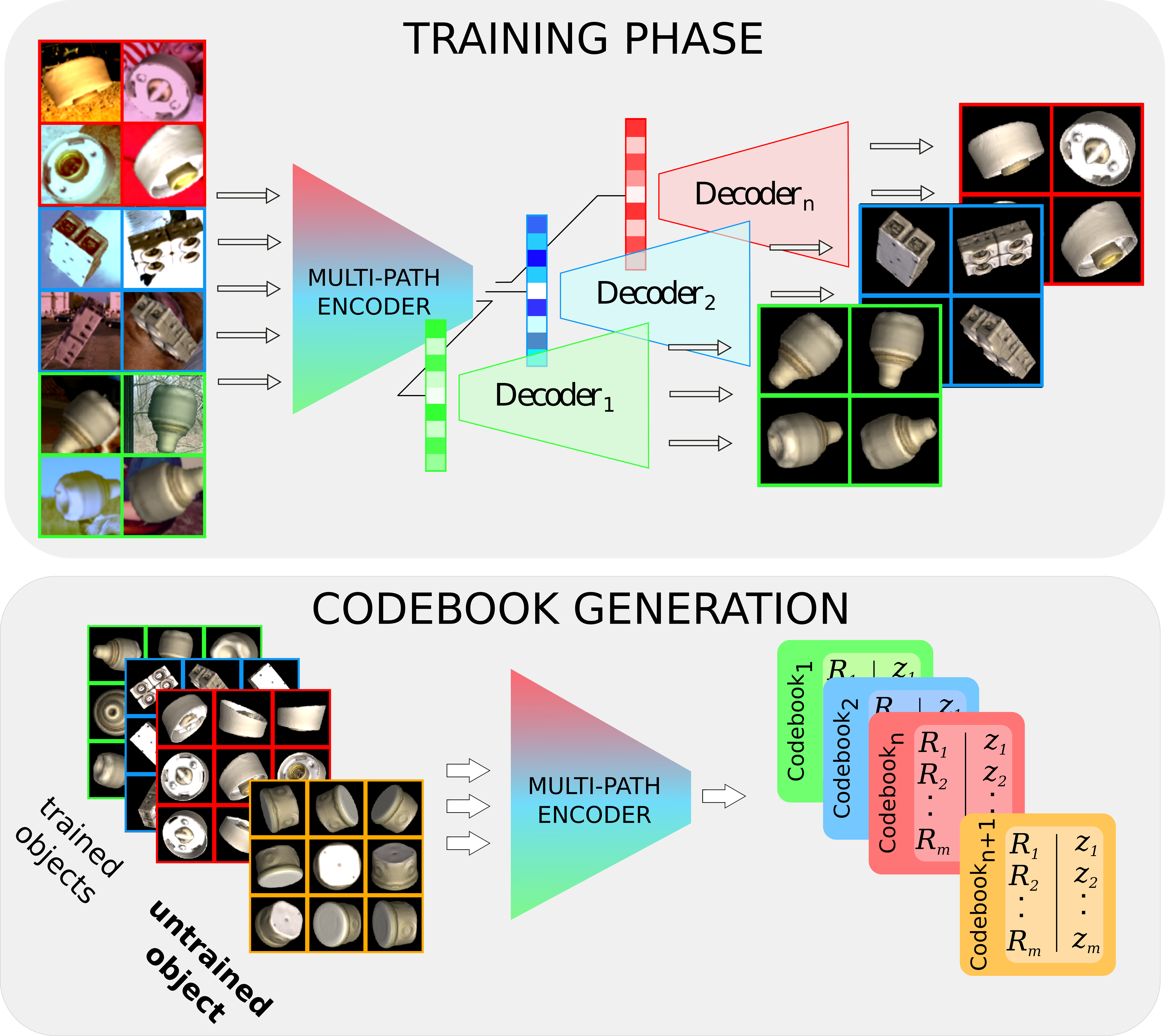}
	\caption{
		Training (top) and setup phase (bottom) of the MP-Encoder. During training one encoder is shared among all objects, while each decoder reconstruct views of a single object. This turns the encoder into a viewpoint-sensitive feature extractor, that generates expressive encodings for multiple trained and even untrained objects.
	}
	\label{fig:front figure}
\end{figure}


Inspired by \glspl{AAE}~\cite{sundermeyer2018implicit} that extract pose representative features on an instance level, we propose a single-encoder-multi-decoder network for jointly estimating the 3D object orientation of multiple objects. While the encoder and latent space are shared among all objects, the decoders are trained to reconstruct views of specific instances (top of Fig.~\ref{fig:front figure}). This \emph{multi-path learning} strategy allows similarly shaped objects to share the same encoding space.
After training, we can generate instance-specific codebooks containing encodings of synthetic object views from all over SO(3). So each entry contains a shape and viewpoint dependent descriptor mapped to an explicit 3D object orientation that can be retrieved at test time. (bottom of Fig.~\ref{fig:front figure})

As we show experimentally, the learned encodings generalize well to the real domain and are expressive enough to relate views from all trained and even untrained objects in a viewpoint sensitive way. 
For a large number of objects the performance of the Multi-Path Encoder (MP-Encoder) does not deteriorate compared to separately trained encoders for each object and even slightly improves encoding quality which leads to new state-of-the-art results in 6-DoF object pose estimation on the T-LESS dataset.

Motivated by this, we also introduce an iterative render-inference scheme based on the learned encodings, which enables relative pose estimation on untrained objects. It resembles the online creation of a local codebook which also helps avoiding SO(3) discretization errors.
We apply this method to iteratively refine poses of untrained instances from ModelNet40 and outperform DeepIM \cite{li2018deepim}, a state-of-the-art approach for RGB-based 6D pose refinement.

\section{Related Work} 

Object pose estimation has been widely studied in the literature,
mainly from the below aspects.

\subsection{Feature-/Template-based approaches} 
Traditionally, pose estimation methods rely on matching algorithms
with local keypoints and descriptors or templates. \Eg, both Hinterstoisser \etal
\cite{hinterstoisser2016going} and Vidal \etal
\cite{vidal20186d} apply so-called \gls{PPF} to match a 3D model into
a given scene outperforming other results in the BOP challenge \cite{hodan2018bop} but having high runtimes per target due to the exhaustive search in the large 6D pose space. While these methods require depth
information, which can be noisy and sensible to environment changes,
Ulrich \etal \cite{ulrich2009cad} use single RGB images and estimate
object poses by matching edges of a rendered 3D model into the image.
In general, the benefit of such approaches using explicit matching is
that no costly training process is required. However, they tend to
require longer execution times. 
Additionally, usually more expert knowledge is required to set up such pipelines, as feature extraction and matching parameters do not generalize well.

\subsection{Learning-based Approaches} 
More recently, a number of learning-based approaches have been
presented with the advantage of reduced computation time and more complex features that can be learned directly from data.
One approach is to classify or regress a representation of the object pose directly \cite{kehl2017ssd, mahendran20173d}. 
Xiang \etal \cite{xiang2017posecnn} propose a CNN architecture with multiple
stages, where the authors define the pose regression loss as a closest object point measure. 
Another line of research makes predictions on sparse or dense 2D-3D keypoint locations, e.g. the 3D bounding box corners \cite{rad2017bb8, tremblay2018deep, tekin2018real} or 3D object coordinates on the surface \cite{Li_2019_ICCV, zakharov2019dpod,Park_2019_ICCV}. The pose is then computed by solving the \gls{PnP} problem.
Apart from occlusions, object and view symmetries cause challenges since they introduce one-to-many mappings from object views to poses. While there are strategies to adapt the loss, network or labels for symmetric objects \cite{wohlhart2015learning, hinterstoisser2012gradient, kehl2017ssd}, dealing with view-dependent or occlusion-induced symmetries is often intractable and negatively affects the training process.
Therefore, a third approach is to learn descriptors of object views that are either conditioned on the pose \cite{balntas2017pose,wohlhart2015learning} or completely implicit \cite{sundermeyer2018implicit}. The advantage of the implicit approach is that symmetric views share the same target representation during trainig. Corresponding explicit poses are automatically grouped and can be retrieved at test time.
These view-centric representations were also shown to work well for 6D tracking \cite{deng2019poserbpf}.

\subsection{Synthetic training data} 
The demand of learning-based methods for large amounts of labeled training data is particularly unfavorable for object pose estimation. Although there are ways
to label object poses semi-automatically \cite{marion2018label}, it
remains a cumbersome, error-prone process. Recent research tackles this by
exploiting synthetic training data \cite{denninger2019blenderproc} which solves this problem but creates a new one through the domain gap between real and synthetic data.

To bridge this gap, Tremblay \etal \cite{tremblay2018deep} use mixed
training data consisting of \textit{photo-realistic rendering} and
\textit{domain randomized} samples.
Since photo-realistic renderings are costly, we follow a less
expensive \gls{DR} strategy \cite{kehl2017ssd,tekin2018real}: 
3D models are rendered in poses 
randomly sampled from SO(3) using OpenGL with random lighting and superimposed onto real images from
data sets such as MS COCO \cite{lin2014microsoft} or Pascal VOC
\cite{everingham2010pascal}.


\subsection{Generalization to Novel Objects} 

Most learning-based methods predict the pose of one \cite{Park_2019_ICCV} or few instances \cite{tremblay2018deep, zakharov2019dpod, tekin2018real, xiang2017posecnn} and have to be retrained for every newly encountered object. However, in domains like service and industrial robotics or augmented reality, it would be beneficial to have a general feature extractor that could produce pose-sensitive features for untrained objects such that testing on a novel object becomes immediately possible. 

When trained on few instances, current pose networks like \cite{tremblay2018deep, tekin2018real} concurrently classify objects which potentially hinders their ability to generalize to untrained objects. Wohlhart \etal \cite{wohlhart2015learning} and Balntas \etal \cite{balntas2017pose} were the first to report qualitative results of deep pose descriptors applied to untrained objects. However, their descriptors are discriminated by both, orientation and object class. So if an untrained object has similar appearance from any viewpoint to one of the trained objects, the corresponding descriptor will get corrupted.
Unlike \cite{wohlhart2015learning,balntas2017pose}, our multi-path training strategy does not separate different object instances in the encoding space and instead allows them to share the same latent features.

Category-level pose estimation \cite{sahin2018category, esteves2018crossdomain, suwajanakorn2018keypoints} can be used to generalize to novel objects from a given category. It assumes that all instances within a category have similar shape and are aligned in a joint coordinate frame. However, these assumptions often do not hold in practice where semantic and geometric similarities often do not coincide. Assigning aligned coordinate frames can be ambiguous because symmetries of instances can vary within a category. Therefore, in this work, we will not explicitly enforce the alignment within semantic categories and instead leave this decision to the self-supervised, appearance-based training.

CNNs trained on large datasets are frequently used to extract low-level features for downstream tasks, e.g. image retrieval~\cite{hao2016best} or clustering \cite{donahue2014decaf}. A naive baseline to predict the 3D orientation of unknown objects would be to compare feature maps of a network trained on a large dataset like ImageNet or COCO. Unsurprisingly this baseline does not work at all because (1) early features are sensitive to translation while the later layers lost geometric information (2) features strongly differ for synthetic and real object views (3) the dimensionality of feature maps is too high to finely discretize SO(3) while reduction techniques like PCA destroy much information.
The pose refinement methods \cite{Manhardt_2018_ECCV,li2018deepim} iteratively predict rotation and translation residuals between an estimated and a target view of an object. The former was shown to generalize to untrained objects of the same category, and the latter even generalizes to objects of new categories. 
These approaches predict an accurate, relative transformation between two object views in a local neighborhood. In contrast, our method is able to yield both, local relative and global 3D orientation estimates.

\section{Method}

We will first briefly describe the \gls{AAE} (Sec. \ref{sec:aae}). Building upon these results, we then propose a novel Multi-Path Encoder-Decoder architecture and training strategy (Sec. \ref{sec:mpencoder}). Next, we will investigate the encoder on its ability to extract pose-sensitive features and examine generalization to novel instances (Sec. \ref{sec:pca}). Different application scenarios depending on the test conditions are discussed (Sec. \ref{sec:generalization}). Finally, an iterative render-inference optimization for pose refinements is presented (Sec. \ref{sec:codebook}).

\subsection{Implicit Object Pose Representations}
\label{sec:aae}
Sundermeyer \etal \cite{sundermeyer2018implicit} have shown that
implicit pose representations can be learned in a self-supervised way using an encoder-decoder
architecture. This so-called \gls{AAE} allows to encode 3D
orientations from arbitrary object views, generalizes from synthetic
training data to various test sensors, and inherently handles symmetric
object views.  

The \gls{AAE} is trained to reconstruct rendered views of a single object. To encode 3D orientation exclusively, the inputs are randomly translated and scaled while the reconstruction target stays untouched. To encode object views from real images, the background of input views are randomized, occlusions at various locations are generated, and various lighting and color augmentations are produced. As a result of this \emph{domain randomization}, the network learns to represent the objects' orientation of real object views implicitly using the latent code $\mathbf{z}$.

Concretely, an input sample $\mathbf{x}\in\mathbb{R}^d$ is randomly augmented by $f(.)$ and mapped by an encoder $\Upsilon$ onto a latent code $\mathbf{z}\in\mathbb{R}^m$ where $m\ll d$. The decoder $\Lambda:\mathbb{R}^m\rightarrow \mathbb{R}^d$ is trained to map the code back to the target $\mathbf{x}$.

\begin{equation}
\hat{\mathbf{x}} = \Lambda(\Upsilon(f(\mathbf{x}))) = \Lambda(\Upsilon(x')) = \Lambda(\mathbf{z})
\end{equation}
Both $\Upsilon$ and $\Lambda$ are neural networks, and their weight
parameters are trained such that the $\ell_2$-loss is minimized, \ie
\begin{equation}
\ell_2( \mathcal{B} ) = \sum_{i\in\mathcal{B}} \|
\mathbf{x}_i - \hat{\mathbf{x}}_i \|_2 =  \sum_{i\in\mathcal{B}} \|
\mathbf{x}_i - \Lambda(\Upsilon(f(\mathbf{x}_i))) \|_2
\label{eq:l2loss}
\end{equation}
where $\mathcal{B}$ contains the indices of input samples of a given batch. After training, the decoder is discarded and latent encodings of object views from all over SO(3) are saved in a codebook together with their corresponding orientations assigned. At test time, a real object crop is encoded and the nearest code(s) in the codebook according to cosine similarity yield the predicted orientation. The rotation can further be corrected for the translational offset as described in \cite{Sundermeyer2019}.

A downside of this formulation is that a new network must be trained for every new object instance. 
When naively training the original \gls{AAE} jointly on several objects, they need to be separated in the latent space so that the decoder is able to reconstruct the correct object.
Even when conditioning the decoder on the object by concatenating a one-hot vector to the encoding, it can only reconstruct few instances and it diminishes the ability of the encoder to encode object orientations. 

\subsection{Multi-Path Encoder-Decoder}
\label{sec:mpencoder}

We propose a simple but effective architecture which, in
combination with our multi-path learning strategy, enables the 3D
orientation estimation of multiple objects (see
Fig. \ref{fig:front figure}). Our architecture consists of a
single encoder $\Upsilon$, an encoding $\mathbf{z} \in
\mathbb{R}^{128}$, and $n$ decoders $\Lambda_j$ with $j = 1,...,n$
where $n$ is the number of different object instances. The
convolutional encoder is fed with the same augmented inputs as an
\gls{AAE} but with heterogeneous batches $\tilde{\mathcal{B}}$ containing multiple objects. The resulting codes are split and each decoder only receives codes that
correspond to a single object instance. The multi-path loss function can
be written as:

\begin{align}
\ell_m(\tilde{\mathcal{B}}) = & \sum_{j=1}^{b}  \sum_{k=1}^{n}
\mathbb{I}(s_j = k) \lVert \mathbf{x}_j - \Lambda_k (\Upsilon
(f(\mathbf{x}_j))) \rVert_2 \nonumber\\ 
=&  \sum_{j=1}^{b} \sum_{k=1}^{n} \mathbb{I}(s_j = k)  \lVert \mathbf{x}_j - \Lambda_k(\mathbf{z}_j) \rVert_2 
\end{align}

where $\mathbb{I}$ is the indicator function used to select the
decoder $\Lambda_k$ that corresponds to the instance $s_j$.
Note that in this setting only the encoder $\Upsilon$ receives
information from the entire mini-batch, while the decoders $\Lambda_j$
backpropagate a loss  $\ell_j$ from a sub-batch.
Since the decoders are only used to learn an efficient encoding, they can be discarded after training, leaving behind a compact encoder model.

In contrast to other approaches
\cite{wohlhart2015learning,balntas2017pose}, where objects are
explicitly separated in the descriptor space, our encoder can learn an
interleaved encoding where general features can be shared across
multiple instances. We consider this ability as the main qualification
to obtain encodings that can represent object orientations from novel,
untrained instances, even when they belong to untrained categories.

\subsection{Principal Component Analysis of Encodings}
\label{sec:pca}
\begin{figure}[t]
	\centering
	\captionsetup{width=1.0\columnwidth}
	\includegraphics[width=1.0\columnwidth]{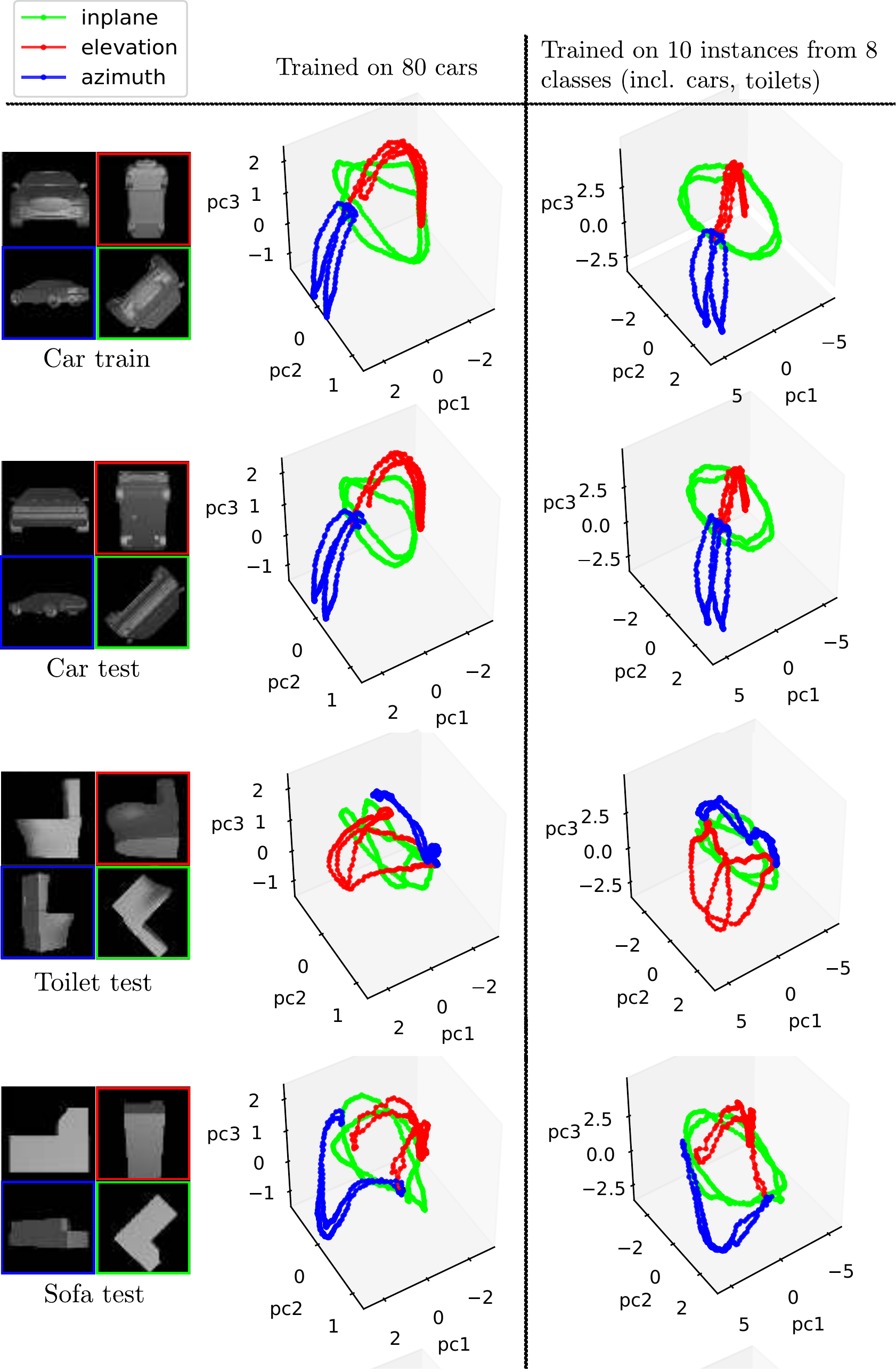}
	\caption{Principal Component Analysis of the learned encodings. Depicted are the encodings of views around elevation (red), azimuth (blue), and in-plane (green). Middle column: Encoder trained only on cars; Right column: Encoder trained on objects from 8 categories}
	\label{fig:pca}
\end{figure}

In order to gain insights into the characteristics of the latent space from trained and untrained objects, we perform an experiment on the ModelNet40 \cite{Wu_2015_CVPR} dataset. We first train the multi-path encoder-decoder on 80 CAD instances that originate from the \textit{car} class. A second model is trained on 10 instances from 8 different classes, namely \textit{airplane, bed, bench, car, chair, piano, sink, toilet}. Training details can be found in the appendix.

After training, we generate 72 viewpoints along a full revolution of azimuth, elevation and in-plane rotation. We record different objects from these viewpoints and feed them into the encoder. From the encoding space $z_i \in \mathbb{R}^{128}$  of all objects we compute the first three principal components and project all encodings onto these directions. The interpolated results can be seen in Fig. \ref{fig:pca}. The top row shows the encodings of a car instance from the training set. The other rows show instances which are not in the training set of neither model, but different sofa and toilet instances were used to train the second model. 

First, it is noticeable that the encodings vary smoothly along each of the rotation axes, even when views of untrained objects are evaluated. It can be seen that the views, while starting at the same point, end up in different sub-manifolds that capture the shape of the encoded objects and their symmetries. For example, the car instance produces closed trajectories with two revolutions because the car shape looks similar at opposite viewpoints. 

Furthermore, it can be seen that the car in the training and test set are encoded similarly by each of the models. This indicates that the encoder as well as the codebook could be reused when predicting orientations of novel cars, even without access to their 3D models.

The remaining encodings of untrained objects can still be clearly separated and thus a meaningful pose descriptor could be extracted without retraining the encoder but simply by creating a codebook from their 3D model.

Apart from an offset, both models learn quite similar encodings, even for the sofa category which the first model has not seen during training. This implies that low-level features are extracted that generalize across various shapes and categories. However, to fulfill the reconstruction task the features still need to be robust against lighting and color augmentations as well as translations that are applied to the inputs. In the next section, we will explore different approaches to utilize these properties.

\subsection{Object Pose Estimation Across Domains}
\label{sec:generalization}
{\setlength{\tabcolsep}{0.8em}
	\begin{table}[t]
		\centering
		\small
		\caption{Different application scenarios for global and iterative pose estimation and the handling of untrained objects}
		\begin{tabular}{cccc}
			& \multicolumn{3}{c}{\textbf{Scenario}} \\
			\cmidrule{1-1}\cmidrule(l){2-3}
			\textbf{Prerequisites}	&  \RNum{1}& \RNum{2} \\
			\cmidrule{1-1}\cmidrule(l){2-3}
			Shape category trained &  \xmark & \cmark \\
			3D Test Model available & \cmark & \xmark \\
			3D Test Model aligned to Train Model& \xmark & \xmark \\
			\cmidrule{1-1}
			\textbf{Available Methods}	& $\Downarrow$&$\Downarrow$ \\
			\cmidrule{1-1}
			Reuse codebook from a trained instance &\xmark& \cmark \\
			Create a new full / sparse codebook & \cmark&\xmark  \\
			Iterative relative pose refinement &\cmark& \xmark \\
			
			
		\end{tabular}
		\label{tab:encoding_test}
	\end{table}
}
After training, the MP-Encoder can create codebooks for all $n$ training instances for pose estimation. (see Sec. \ref{sec:aae})

On top of that, it can deal with scenarios inlcuding untrained objects which are depicted in Table \ref{tab:encoding_test}. Here, the trained encoder model is used as a fixed pose-sensitive feature extractor. Available methods depend on the characteristics and given information on the considered test objects. 

If a 3D model of the untrained object is available, it is usually preferable to create a new codebook and estimate the 3D orientation from it (I). If no 3D model is available, but we have codebooks of trained instances from a category with coinciding shapes, we could also reuse these codebooks directly for the test instance (II). However, as we do not explicitly align models of a category in a single coordinate frame, the extracted representation will be more dependent on the shape of an object than on any semantics.

Given a 3D model, we can also use our method for iterative pose refinement at test time. This allows us to refine results from sparser codebooks or to perform local, relative pose estimation directly without any codebook.

\begin{algorithm}[t]
	\small
	\SetAlgoLined
	\KwIn{encoder $\Upsilon$, init pose $\mathbf{q_{init}, t_{init}}$, target view $\mathbf{x}_\ast$}
	$\mathbf{z}_\ast \leftarrow \Upsilon(\mathbf{x}_\ast)$ \\
	$\mathbf{q_{est}} \leftarrow \mathbf{q_{init}}$ \\
	$\mathbf{t_{est}} \leftarrow \mathbf{t_{init}}$ \\
	\For{$k=0\dots2$}{
		\For{$j=0\dots3$}{
			\For{$i=0\dots40 - 10k$}{
				$\alpha \sim \mathcal{N}(0,\,\frac{\sigma^{2}}{j+1})$ \\
				$\mathbf{v} \sim \mathcal{N}_3(\boldsymbol{0},\,I)$ \\
				$\Delta \mathbf{q}  \leftarrow quat(\frac{\mathbf{v}}{\lVert \mathbf{v} \rVert},\alpha)$ \\
				$\mathbf{q_{i}} \leftarrow \Delta \mathbf{q} \, \mathbf{q_{est}}$ \\
				$\mathbf{x}_{i} \leftarrow render(\mathbf{q_{i}}, \mathbf{t_{est}})$ \\
				$\mathbf{z}_{i} \leftarrow \Upsilon(\mathbf{x}_{i})$ \\
			}
			$k \leftarrow \underset{i}{\arg \max} \frac{\mathbf{z}_i \;\mathbf{z}_\ast}{\lVert \mathbf{z}_i \rVert \lVert \mathbf{z}_\ast \rVert}$ \\
			$\mathbf{q_{est}} \leftarrow  \mathbf{q_k}$
		}
		$\mathbf{x_{est}} = render(\mathbf{q_{est}}, \mathbf{t_{est}})$ \\
		$\Delta \mathbf{t} = multiScaleEdgeMatching(\mathbf{x_{*}}, \mathbf{x_{est}})$\\
		$\mathbf{t_{est}} = \mathbf{t_{est}} + \Delta \mathbf{t} $ \\
	}
	\KwResult{$\mathbf{q_{est}}, \mathbf{t_{est}}$}
	\caption{Iterative 6D Pose Refinement}
	\label{pseudo}
\end{algorithm}
\begin{table*}[t]
	\centering
	\small
	
	\caption{Pose estimation performance of the MP-Encoder trained on objects 1-18 on the complete T-LESS primesense test set (RGB-only) compared to 30 single object \glspl{AAE} \cite{sundermeyer2018implicit} trained on all objects. We measure recall under the \gls{VSD} metric. The right column shows the performance of a model trained only on 30 instances of the ModelNet40 dataset. We use ground truth bounding boxes (top) and ground truth masks (bottom) in this experiment. It can be observed that a single MP-Encoder can reach similar performance on unknown objects if segmentation masks are given.
	}
	\bgroup
	\def\arraystretch{1.2}
	\begin{tabular}{c|c|c|c|c}
		\toprule
		&Mean & 30 separate AAE & \multicolumn{2}{c}{Single Multi-Path Encoder trained on}\\
		&VSD recall &encoders \cite{sundermeyer2018implicit} & T-Less Objects 1-18 &
		30 Instances ModelNet40 \\
		\hline
		\multirow{3}{*}{\STAB{\rotatebox[origin=c]{90}{+gt bbox}}}&Objects 1-18&\textbf{35.60} &\textbf{35.25}& 27.64\\
		&Objects 19-30 & \textbf{42.45}&33.17&34.57\\
		&Total  &\textbf{38.34} &34.42&30.41\\
		\hline
		\hline
		\multirow{3}{*}{\STAB{\rotatebox[origin=c]{90}{+gt mask}}}&Objects 1-18 & 38.98 &\textbf{43.17}& 35.61\\
		&Objects 19-30 & \textbf{45.33} & 43.33 & 42.59\\
		&Total &41.52&\textbf{43.24}&38.40\\
		
	\end{tabular}
	\label{tab:gen_tless}
	\egroup
\end{table*}
\subsection{Iterative Refinement of Latent Codes \label{sec:codebook}}


Our method for iterative pose refinement is outlined in Alg. \ref{pseudo}. We start with an initial pose estimate and a target view. Next, we render a batch of 3D model views at small rotational perturbations from the initial pose and insert the whole batch into the encoder. The code with the highest cosine similarity to the target view encoding determines the new rotation estimate. We sample again with a smaller perturbation radius. The inner loop of this random optimization scheme, consisting of rendering and inference, can be efficiently parallelized. 

Rotation and translation are alternately optimized three times in our experiments. As the MP-Encoder is trained to be invariant against translations, we can first optimize for the more challenging out-of-plane rotation neglecting translation. Afterwards, the rotation is usually roughly aligned to the target and we use a simple edge-based, multi-scale template matching method based on OpenCV \cite{opencv_library} to determine the translational offset. 
The advantage of sampling in SO(3) vs. sampling in the latent space is that (1) the SO(3) space has lower dimensionality and (2) we only search valid orientations.

Apart from relative pose estimation with a novel 3D model without a codebook, the refinement also allows to trade-off between set-up time, inference time, memory resources and accuracy. The full $92232 \times 128$ codebook creation takes $\sim5$ minutes per object on a modern GPU and $45$MB space. Inference only takes $\sim6$ms on a modern GPU while the 6D pose refinement in the presented configuration takes approximately 1 second. To further speed this up, the random search could be replaced by more sophisticated black-box optimization tools such as CMA-ES \cite{hansen2019pycma}.
Depending on the application, i.e. number of objects, available resources and constraints on the set-up time, global initialization with sparse codebooks combined with local pose refinement could be a viable option.

\subsection{6DoF Object Detection Pipeline}
\label{sec:6d}
Our full RGB-based 6DoF object detection pipeline consists of a MaskRCNN with ResNet50 backbone \cite{he2017mask}, the MP-Encoder for 3D orientation estimation and a projective distance estimate \cite{sundermeyer2018implicit} for translation estimation. To compare against other approaches that use depth data, we also report results with a point-to-plane ICP \cite{chen1992object, zhang1994iterative} 
refinement step. Especially in the presence of severe occlusions the RGB-based  projective distance estimate does not produce distances accurate enough to fulfill the strict VSD metric. On the other hand, an accurate initialization is crucial for ICP refinement to work well.

For the MaskRCNN we generate training data by pasting object recordings from the T-LESS training set at random translation, scale and in-plane rotation on random background images. Thereby, we produce 80000 images with MS COCO \cite{lin2014microsoft} background, 40000 images with black background and 40000 images with random texture backgrounds \cite{cimpoi14describing}. We apply several standard online color augmentations like hue, brightness, blur and contrast. Our MaskRCNN reaches an mAP@0.5 performance of 0.68 (bbox detection) and 0.67 (segmentation). Both, MaskRCNN and the MP-Encoder can process multiple objects at once and thus achieve a runtime that stays nearly constant for a large number of objects. 
\section{Evaluation}

{\setlength{\tabcolsep}{0.2em}  
	\begin{table*}
		\centering
		\small
		
		\caption{Evaluation of our full 6D Object Detection pipeline with MaskRCNN + Multi-Path Encoder + optional ICP. We report the mean VSD recall following the SIXD challenge / BOP benchmark \cite{hodan2018bop} on the T-LESS Primesense test set. See the appendix for object-wise results. A single MP-encoder model outperforms the result of 30 instance-specific AAEs.}
		\label{tab:full_pipeline}
		
		\bgroup
		\def\arraystretch{1.0}
		\begin{tabular}{c|c|ccc|ccc|cc}
			
			\toprule
			&Template matching& \multicolumn{3}{c|}{\gls{PPF} based} & \multicolumn{5}{|c}{Learning-based}\\
			\toprule
			&Hodan-15 & Vidal-18 & Drost-10 & Drost-10-edge& Brachmann-16  & Kehl-16 &\textbf{OURS}&Sundermeyer-18& \textbf{OURS}\\
			& Depth & Depth +ICP & Depth & Depth + RGB &RGB-D & RGB-D + ICP & RGB + ICP & \multicolumn{2}{c}{RGB only}\\
			\midrule
			Average  & 63.18&  66.51&56.81  	 &	67.5  	 & 17.84 &	24.6&   \textbf{69.53}	&19.26&\textbf{20.53} \\
			\midrule
			Time (s)  & 13.5 &  4.7  	 &	2.3		 & 21.5	      	 &		4.4  &	1.8	& \textbf{0.8} & \textbf{0.1} & 0.2 
		\end{tabular}
		\egroup
	\end{table*}
}


We focus our analysis on two benchmarks: ModelNet40 \cite{Wu_2015_CVPR} and the T-LESS dataset \cite{hodan2017tless}. \footnote{Further results can be expected at the BOP challenge 2020 \cite{Hodan_2018_ECCV}}

\subsection{Metrics}
\label{metrics}
In ModelNet40 we use the absolute angular error
\begin{equation}
e_R = \arccos\,\Big(tr(\hat{R}^T \, R-I)\,/\,2\Big)
\end{equation} 
as well as the ADD metric \cite{hinterstoisser2012model} at an average distance threshold of $0.1 \times$ object diameter for the model points $\mathcal{M}$
\begin{equation}
ADD = \frac{1}{m} \sum_{x \in \mathcal{M}}||(Rx + t) - (\hat{R}x+\hat{t})||
\end{equation} 
and the 2D projection metric at an average 5px threshold
\begin{equation}
Proj2D = \frac{1}{m} \sum_{x \in \mathcal{M}}||K(Rx + t) - K(\hat{R}x+\hat{t})||
\end{equation} 
In the T-LESS experiments we use the \gls{VSD} \cite{hodan2016evaluation} since here the above metrics are quite meaningless as a result of various object and view symmetries. The \gls{VSD} is an ambiguity-invariant pose error metric that is computed from the distance between the estimated and ground truth visible object surfaces. As in the SIXD challenge 2017 \cite{sixd} and the BOP benchmark 2018  \cite{Hodan_2018_ECCV}, we report the recall of 6D object poses at $err_{vsd} < 0.3$ with tolerance $\tau = 20mm$ and $>10\%$ object visibility. 

%

\subsection{Generalization Capabilities of the MP-Encoder}

We first investigate the joint encoding performance and generalization capabilities of the MP-Encoder in isolation, i.e. with ground truth detections and masks. Therefore, we compare one MP-Encoder against 30 separately trained \gls{AAE} models on all scenes of the T-LESS Primesense test set (Table \ref{tab:gen_tless}). Equivalent encoder and decoder architectures are used. Furthermore, the MP-Encoder is only trained on the first 18 3D object reconstructions of the T-LESS dataset to show the generalization capabilities on untrained objects 19-30. On objects 1-18 which the MP-Encoder has seen during training, the results using ground truth bounding boxes are close to the \gls{AAE} results. Looking at the next row, the performance on the untrained objects 19-30 is significantly worse compared to the single \glspl{AAE}. We suppose that the MP-Encoder has difficulties to extract unknown objects from the background. This hypothesis is strongly supported by the results with ground truth masks (bottom) where the MP-Encoder even outperforms the \glspl{AAE}. Even for the untrained objects 19-30 the performance gap to the AAEs that were trained on these objects is quite small. One possible explanation is that a feature extractor trained on multiple objects learns more versatile and thus more robust features.

The last column of Table \ref{tab:gen_tless} depicts the surprisingly good generalization from ModelNet40 to the T-LESS dataset. Here, the MP-Encoder is specifically trained on 30 texture-free CAD models from the 8 categories \textit{airplane, bed, bench, car, chair, piano, sink, toilet}. Codebooks are created for all T-LESS objects and it is tested on the same real sensor recordings. These results underline that with the multi-path training combined with an input randomization strategy, we can learn to extract orientation variant features that generalize well across greatly differing domains.
\begin{figure}[t]
	\centering
	\captionsetup{width=\linewidth}
	\includegraphics[width=\linewidth]{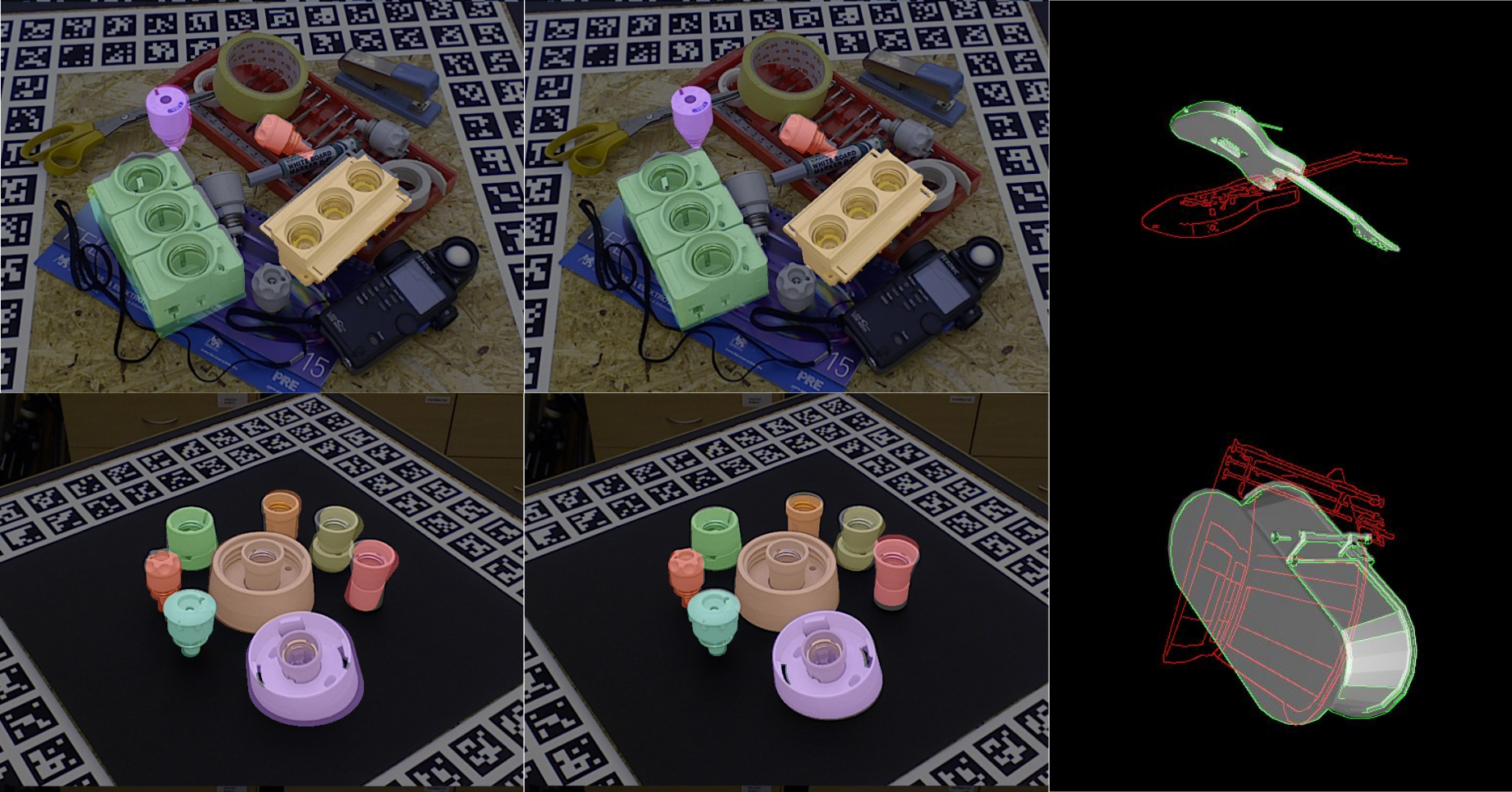}
	\caption{Left: Qualitative 6D Object Detection results of the RGB-based pipeline on T-LESS Primesense scenes. Only 18 of 30 objects are used to train the Multi-Path encoder; Middle: ICP-refined results; Right: Relative pose refinement of instances from the untrained categories guitar and bathtub. Red is the initial pose, green depicts the refined pose.}
	\label{fig:results}
\end{figure}%
{\setlength{\tabcolsep}{0.4em}  
	\begin{table*}
		\centering
		\small	
		\caption{Relative Pose Refinement from up to $45^o$ and $\Delta t = (10,10,50) [mm]$ perturbations on untrained instances of seen (top) and unseen (bottom) object categories of the ModelNet40 dataset. We measure the recall at the $5cm, 5^o$ threshold, the ADD at 0.1d (object diameter) metric and the Proj2D at 5px threshold.
		}
		\bgroup
		\def\arraystretch{1.07}
		\begin{tabular}{c|l|cccc|cccc|cccc}                                                                                            
			\toprule                                                                                                            
			&metric &   \multicolumn{4}{c|}{$(5^o, 5cm)$} & \multicolumn{4}{c|}{6D Pose (ADD)} &  \multicolumn{4}{c}{Proj2D (5px)} \\   
			
			&\multirow{2}{0pt}{method} & \multicolumn{2}{c}{DeepIm \cite{li2018deepim}} & \multicolumn{2}{c|}{Ours} &   \multicolumn{2}{c}{DeepIm \cite{li2018deepim}} & \multicolumn{2}{c|}{Ours} & \multicolumn{2}{c}{DeepIm \cite{li2018deepim}} & \multicolumn{2}{c}{Ours} \\                                                                                
			&&  init &  refined &  init &  refined &  init &  refined & init &  refined &  init &  refined &  init &  refined \\       
			\midrule
			\multirow{4}{*}{\STAB{\rotatebox[origin=c]{90}{novel instance}}}&airplane &    0.8 & 68.9 &      0.9 &        \textbf{96.9} &       25.7 & 94.7 &           33.4 &             \textbf{97.9} &          0.4  & 87.3 &          0.1 &              \textbf{97.4} \\
			&car &         1.0 & 81.5 &   0.4 &        \textbf{96.4} &       10.8  & 90.7 &         13.4 &             \textbf{98.5} &             0.2 & 83.9 &       0.1 &              \textbf{94.0} \\
			&chair &        1.0 & 87.6 &     0.3 &        \textbf{96.4} &       14.6 & 97.4 &           16.3 &             \textbf{98.3} &           1.5  &88.6 &        0.0 &              \textbf{94.6} \\
			\cmidrule{2-14}
			&Mean &        0.9 & 79.3 & 0.5 &        \textbf{96.6} &        17.0&  94.3 &         21.0 &             \textbf{98.2} &              0.7&  86.6 &      0.1 &              \textbf{95.3} \\
			
			\midrule                                                                             
			\midrule                                                                                                            
			\multirow{8}{*}{\STAB{\rotatebox[origin=c]{90}{novel category}}}&bathtub &        0.9 & 71.6 &      0.7 &       \textbf{85.5} &         11.9 & 88.6 &          15.4 &             \textbf{91.5} &         0.2 & 73.4 &           0.1 &              \textbf{80.6} \\
			&bookshelf &      1.2 & 39.2 &        0.7 &     \textbf{81.9} &       9.2 & 76.4 &            13.7 &             \textbf{85.1} &        0.1 & 51.3 &            0.0 &              \textbf{76.3} \\
			&guitar &         1.2 & 50.4 &     0.5 &        \textbf{69.2} &          9.6 & 69.6 &         13.1 &             \textbf{80.5} &           0.2 & 77.1 &         0.3 &              \textbf{80.1} \\
			&range\_hood &    1.0 & 69.8 &          0.5 &   \textbf{91.0} &     11.2 & 89.6 &              14.1 &             \textbf{95.0} &     0.0 & 70.6 &               0.0 &              \textbf{83.9} \\
			&sofa &           1.2 & 82.7 &   0.6 &          \textbf{91.3} &            9.0 & 89.5 &       12.2 &             \textbf{95.8} &             0.1 & \textbf{94.2} &       0.0 &      86.5 \\
			&wardrobe &       1.4 & 62.7 &       0.7 &      \textbf{88.7} &        12.5 & 79.4 &           14.8 &             \textbf{92.1} &        0.2 & 70.0 &            0.0 &              \textbf{81.1} \\
			&tv\_stand &      1.2 & 73.6 &        0.6 &     \textbf{85.9} &       8.8 & \textbf{92.1} &   10.5 &                      90.9 &        0.2 & 76.6 &            0.1 &              \textbf{82.5} \\
			\cmidrule{2-14}
			& Mean &           1.2 & 64.3 & 0.6 &        \textbf{84.8}    & 10.3 &  83.6    &       13.4 &             \textbf{90.1} &       0.1 &     73.3 &    0.1 &            \textbf{81.6} \\
			
			\bottomrule
		\end{tabular}
		
		\label{tab:relative_pose}
		\egroup
	\end{table*}
}
\subsection{6D Object Detection results}
Next, we evaluate our full 6D pose estimation pipeline on the T-LESS dataset which is a particularly challenging 6D object detection benchmark containing texture-less, symmetric objects as well as clutter and severe occlusions. 
Table \ref{tab:full_pipeline} presents results using the strict vsd metric (Sec. \ref{sec:6d}). We achieve state-of-the-art results on T-LESS at much lower runtimes than previous methods, both in the RGB domain and also in the depth domain when initializing an ICP with our RGB-based pose estimate.
\begin{figure}[t]
	\centering
	\captionsetup{width=0.9\columnwidth}
	\includegraphics[width=0.35\columnwidth]{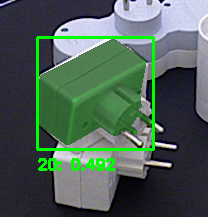}
	\caption{Failure case: MaskRCNN predicts the wrong class 20 instead of 19 (the object below). Since the shape is quite similar (except scale) the codebook of object 20 still gives a reasonable pose estimate.}
	\label{fig:failure}
\end{figure}%
Although the gains are marginal, the results are significant because we only train a single encoder on the whole dataset and still obtain state-of-the-art results. This makes our method scalable, particularly considering that no real pose-annotated data is used for training. Fig. \ref{fig:results} (left) shows qualitative examples of the full 6D Object Detection pipeline. Previously, pose estimation of texture-less objects has been dominated by pure depth-based geometric matching approaches that often possess high run-times and do not scale very well with the number of considered objects. Fig. \ref{fig:failure} shows a failure case which underlines that strict instance-wise pose estimation may not be suitable for real world applications where objects often differ only in details. 

\subsection{Iterative Refinement on Untrained Objects}
\label{sec:iter_unseen}

In our final experiment, we evaluate the MP-Encoder on the task of iterative refinement of poses on untrained instances from seen and unseen categories from ModelNet40. 

We follow the evaluation protocol of DeepIm \cite{li2018deepim} where the relative pose between two object views of an untrained instance is sought. The target view is rendered at constant translation $t = (0,0,500)$ (mm) and random rotation $R \sim SO(3)$. Then we render another object view with that pose perturbed by the angles $\beta_{x/y/z} \sim \mathcal{N}(0,\,(15^o)^2)$ sampled for each rotation axis and a translational offset $\Delta x \sim \mathcal{N}(0, 10^2)$, $\Delta y \sim \mathcal{N}(0, 10^2)$, $\Delta z \sim \mathcal{N}(0, 50^2)$ (mm). If the total angular perturbation is more than $45^o$, it is resampled.

We train category specific MP-Encoders on 80 instances of the airplane, car and chair class and predict the relative pose on novel instances. We also train another MP-Encoder on 80 instances from 8 different categories \footnote{\textit{airplane, bed, bench, car, chair, piano, sink, toilet}} to obtain a general viewpoint-sensitive feature extractor which we test on instances from novel categories. To retain the relative pose to the target view we use our random optimization scheme described in Alg. \ref{pseudo} with an initial $\sigma = 45^o$. 

We compare our approach against DeepIm \cite{li2018deepim} that also minimizes the relative pose between an initial and a target view through an iterative rendering inference cycle. The results in Table \ref{tab:relative_pose} demonstrate superior performance of our approach on both, seen and unseen categories. Figure \ref{fig:results} on the right shows qualitative results of the refinement on unseen categories.

\section{Conclusion}

In this paper we presented a novel approach for estimating poses of multiple trained and untrained objects from a single encoder model.
In contrast to other methods, training on multiple objects does not degrade performance and instead leads to state-of-the-art results on the texture-less, symmetric objects of the T-LESS dataset. The same MP-Encoder architecture is also used for iterative pose refinement on untrained objects outperforming previous approaches on ModelNet40.

The ability of this pose estimator to generalize across objects from different categories, datasets and image domains indicates that low-level viewpoint-sensitive features are shared across various domains. Higher-level features for discriminating semantics require viewpoint-invariance and usually generalize less well. Therefore, our results suggest that these two tasks should be learned separately.

We believe that this is a step towards coping with the large number of instances in industrial settings where 3D models are often available and service robotics where categories often appear repeatedly and interactions with novel objects should be immediately possible.


\newpage

{\small
\bibliographystyle{ieee_fullname}
\bibliography{refs}
}

\newpage

\appendix

\section{Appendix}
In the following we provide implementation details and show qualitative results. The code will be open sourced upon publication.

\subsection{Architecture}\label{subsec:arch_and_training}
We closely follow \cite{sundermeyer2018implicit} to design our Multi-Path (MP) encoder and decoders.

In our T-LESS experiments the encoder is symmetric to each of the decoders. The encoder consists of four convolutional layers with kernel size = 5, stride=2, ReLU activation and $\{128,256,512,512\}$ filters respectively. It follows, a shared fully connected layer with $W_0 \in \mathcal{R}^{32768x128}$. The resulting codes are propagated to the corresponding decoders via $j \in 1,..,n$ fully connected layers $W_j \in \mathcal{R}^{128x32768}$. After reshaping, each decoder has four consecutive convolutional layers with kernel size = 5, stride=1, ReLU activation and $\{512,512,256,128\}$ filters. Nearest neighbor upsampling is performed after each convolutional layer. The final reconstruction output is preceded by a sigmoid activation.

We initially expected that an encoder with up to hundred corresponding decoders would require a much larger architecture to be able to adequately encode the views of all training objects. Therefore, we first adopted the  \cite{chen2018encoder}, with a Resnet101 backbone and a spatial pyramid pooling (SPP) layer. Although, the total training loss can be reduced with such a network, the downstream performance on encoding object poses of both, trained and untrained objects, does not increase compared to a more shallow encoder/decoder with 4-5 plain convolutional layers as used by \cite{sundermeyer2018implicit}. As usual, multiple explanations are plausible. Shallow features could contain all necessary information for class-agnostic pose estimation. The lost symmetry in the encoder-decoder architecture could make an invertible compression more difficult to learn. A deeper encoder could simply overfit more to synthetic and generalize less to real data. As architecture search and explanation is not our main goal here, we simply stick with the more efficient shallow CNN.


\begin{figure}[h]
	\centering
	\captionsetup{width=1.0\columnwidth}
	\includegraphics[width=1.0\columnwidth]{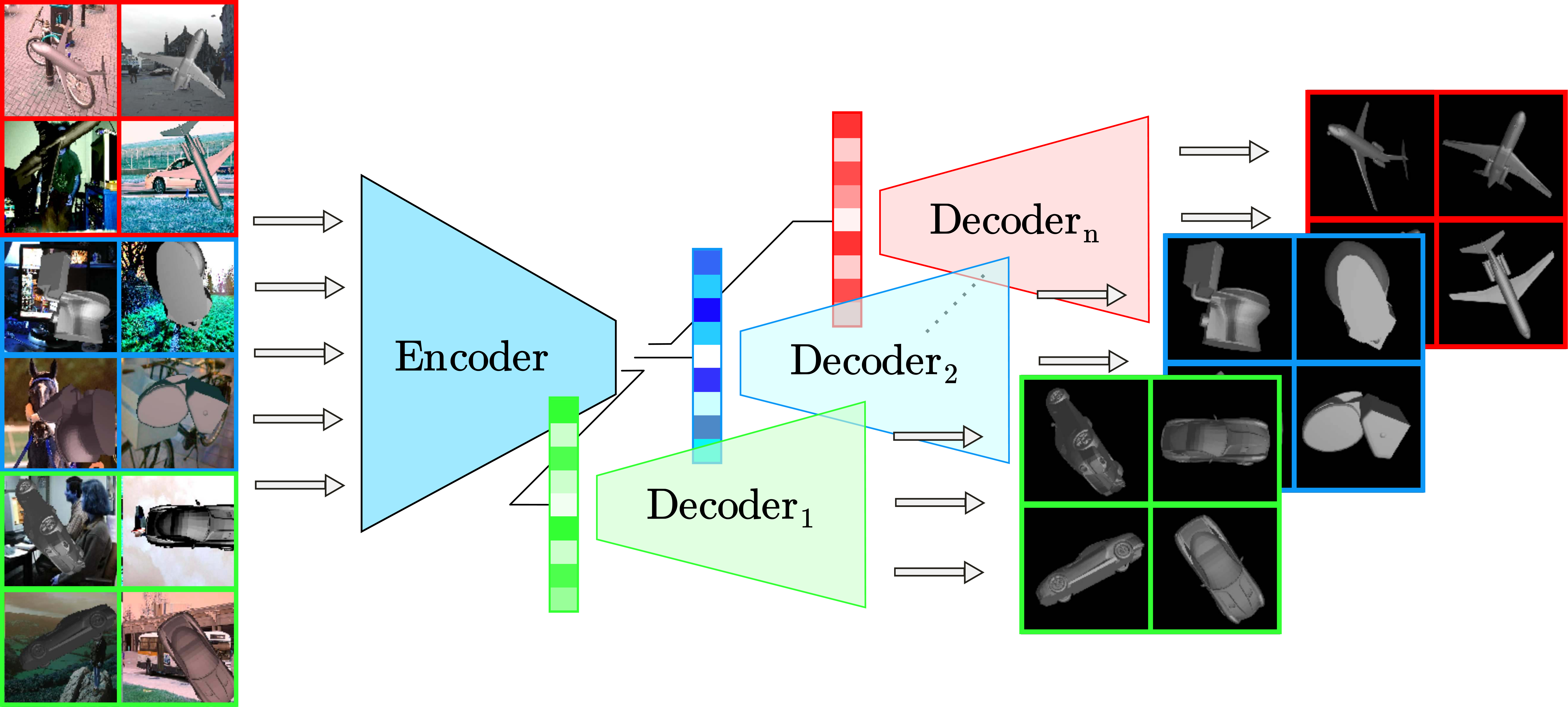}
	\caption{Our multi-path training process. {\bf Left:} Joint augmented input batch with uniformly sampled SO(3) views of $n$ object instances; {\bf Right:} Individual reconstruction targets $\pmb x_j$
	          for each decoder.}
	\label{fig:training_process}
\end{figure}


\subsection{Training Details}
It is possible to train our architecture with up to 80 decoders on a single GPU. In this case, the encoder receives a joint batch size of 80, while each decoder has a batch size = 1. For single-GPU training on T-LESS with 18 decoders, we use a batch size = 4. We train for 300.000 iterations which takes about 48 hours on a single GPU. Multiple GPUs almost linearly speed up training due to the inherent parallelism in the architecture. Therefore, the batch and the decoders are divided into equal parts on all available GPUs. For the ModelNet experiments where we trained on 80 object instances, we used 4 GPUs with a total encoder batch size of 240 and decoder batch size of 4. Higher decoder batch sizes also stabilize the training.

We use Xavier initialization and the Adam \cite{kingma2014adam} optimizer with a learning rate of $10^{-4} \times b_{dec}$, with $b_{dec}$ = decoder batch size. In our experiments we did not have to assign individual learning rates to the encoder and decoder but this could potentially accelerate the training.

\subsection{Synthetic Data Generation}

For each instance we render 8000 object views sampled randomly from SO(3) at a constant distance of 700mm. The resulting images are quadratically cropped and resized to $128 \times 128 \times 3$.  

%

\subsection{Augmentation Parameters}
All geometric and color input augmentations besides the random lighting are applied online during training at uniform random strength (see Table \ref{tab:aug_strong_col}). As background images we use Pascal VOC2012 \cite{pascalvoc2012}. One could argue that random background images during training are redundant where we use MaskRCNN. Training without backgrounds works, but in our experiments it does help with the generalization from synthetic to real data and is also beneficial when the MaskRCNN does not output perfect masks.

\begin{table}[h]
	\footnotesize
	\centering
	\captionsetup{width=0.9\columnwidth}
	\caption{Augmentation Parameters; Scale and translation is in relation to image shape}
	\begin{tabular}{cc|cc}
		\toprule
		color & 50\% chance &\multicolumn{2}{c}{light (random position) } \\
		& (30\% per channel) & \multicolumn{2}{c}{\& geometric}\\
		\midrule
		add & $\mathcal{U}(-0.1,0.1)$ & ambient &$0.4$ \\
		contrast & $\mathcal{U}(0.4,2.3)$ &diffuse &$\mathcal{U}(0.7,0.9)$\\
		multiply & $\mathcal{U}(0.6,1.4)$ & specular&$\mathcal{U}(0.2,0.4)$ \\
		invert &  & scale &$\mathcal{U}(0.8,1.2)$\\
		gaussian blur & $\sigma \sim \mathcal{U}(0.0,1.2)$ & translation & $\mathcal{U}(-0.15,0.15)$\\
	\end{tabular}
	\label{tab:aug_strong_col}
\end{table}

 \subsection{ICP Refinement}
 \label{sec:icp}
 Optionally, the estimate is refined on depth data using a point-to-plane \gls{ICP} approach with adaptive thresholding of correspondences based on \cite{chen1992object, zhang1994iterative}. The refinement is first applied in direction of the vector pointing from camera to the object where most of the RGB-based pose estimation errors stem from and then on the full 6D pose.
  
 {\setlength{\tabcolsep}{0.2em}  
 	\begin{table*}[t]
 		\centering
 		\footnotesize
 		\captionsetup{width=0.93\textwidth,justification = raggedright}
 		\caption{Evaluation of the full 6D Object Detection pipeline with MaskRCNN + Multi-Path Encoder + optional ICP. We report the mean VSD recall following the SIXD Challenge \cite{sixd} on the T-LESS Primesense test set.}
 		
 		\bgroup
 		\def\arraystretch{1.0}
 		\begin{tabular}{c|c|ccc|ccc|cc}
 			
 			\toprule
 			&Template matching& \multicolumn{3}{c|}{\gls{PPF} based} & \multicolumn{5}{|c}{Learning-based}\\
 			\toprule
 			object &Hodan-15 & Vidal-18 & Drost-10 & Drost-10-edge & Kehl-16 &\textbf{OURS}& Brachmann-16 &Sundermeyer-18& \textbf{OURS}\\
 			& Depth & Depth +ICP & Depth & Depth & RGB-D + ICP & RGB + ICP & RGB-D&\multicolumn{2}{c}{RGB}\\
 			\midrule
            1	 & 66		 & 43	&   34	 & 53		 	   			  & 7 	& 73.61 & 8 &9.48 & 5.56   \\ 
			2	 & 67		 & 46	&	 46	 &44		 	    		  &10	& 66.40 &10 &13.24 &   10.22 \\
			3	 & 72		 & 68	&   63	 & 61		 	   			  & 18	& 87.24 & 21&12.78 &	14.74 \\
			4	 & 72		 & 65	&   63	 & 67		 	   			  & 24	& 82.91 & 4 &6.66 & 	6.23  \\
			5	 & 61		 & 69	&   68	 & 71		 	   			  & 23	& 86.16 & 46&36.19 &37.53 \\
			6	 & 60		 & 71	&   64	 & 73		 	   			  & 10	& 92.79 & 19&20.64 & 30.36 \\
			7	 & 52		 & 76	&   54	 & 75		 	   			  & 0 	& 80.83 & 52&17.41 &14.62 \\
			8	 & 61		 & 76	&   48	 & 89		 	   			  & 2 	& 81.32 & 22&21.72 &10.73 \\
			9	 & 86		 & 92	&   59	 & 92		 	   			  & 11	& 85.15 & 12&39.98 &19.43 \\
			10	 & 72		 & 69	& 	 54	 & 72		 	   			  & 17	& 82.31 & 7 &13.37 &32.75 \\
			11	 & 56		 & 68	& 	 51	 & 64		 	   			  & 5 	& 72.60 & 3 &7.78 & 20.34 \\
			12	 & 55		 & 84	& 	 69	 & 81		 	   			  & 1 	& 68.80 & 3 &9.54 &29.53 \\
			13	 & 54		 & 55	& 	 43	 & 53		 	   			  & 0 	& 53.37 & 0 &4.56 &12.41 \\
			14	 & 21		 & 47	& 	 45	 & 46		 	   			  & 9 	& 50.54 & 0 & 5.36 &21.30 \\
			15	 & 59		 & 54	& 	 53	 & 55		 	   			  & 12	& 45.25 & 0 & 27.11 &20.82 \\
			16	 & 81		 & 85	& 	 80	 & 85		 	   			  & 56	& 82.32 & 5 &22.04 &33.20 \\
			17	 & 81		 & 82	& 	 79	 & 88		 	   			  & 52	& 72.27 & 3 &66.33 &39.88 \\
			18	 & 79		 & 79	& 	 68	 & 78		 	   			  & 22	& 80.60 & 54&14.91 &14.16 \\
			19	 & 59		 & 57	&   53 	 &	55		 	   		  	  &	35	& 48.17 & 38&23.03 &	9.24  \\
			20	 & 27		 & 43	&   35 	 &	47		 	   		  	  &	5 	& 25.74 & 1 &5.35 & 	1.72  \\
			21	 & 57		 & 62	&   60 	 &	55		 	   		  	  &	26	& 47.53 & 39&19.82 &	11.48 \\
			22	 & 50		 & 69	&   61 	 &	56		 	   		  	  &	27	& 50.27 & 19&20.25 &	8.30  \\
			23	 & 74		 & 85	&   81 	 &	84		 	   		  	  &	71	& 46.99 & 61&19.15 & 	2.39  \\
			24	 & 59		 & 66	&   57 	 &	59		 	   		  	  &	36	& 78.20 & 1 &4.54 & 	8.66  \\
			25	 & 47		 & 43	&   28 	 &	47		 	   		  	  &	28	& 50.00 & 16&19.07 &	22.52 \\
			26	 & 72		 & 58	&   51 	 &	69		 	   		  	  &	51	& 64.61 & 27&12.92 &	30.12 \\
			27	 & 45		 & 62	&   32 	 &	61		 	   		  	  &	34	& 73.09 & 17&22.37 & 	23.61 \\
			28	 & 73		 & 69	&   60 	 &	80		 	   		  	  &	54	& 86.14 & 13&24.00 &	27.42 \\
			29	 & 74		 & 69	&   81 	 &	84		 	   		  	  &	86	& 77.66 & 6 &27.66 &	40.68 \\
			30	 & 85		 & 85	&   71 	 &	89		 	   		  	  &	69	& 92.96 & 5 &30.53 &	56.08 \\
 			\midrule
 			Average  & \textbf{63.18}&  66.51&56.81  	 &	\textbf{67.5}  	  &	24.6&   \textbf{69.53}	& 17.84&19.26&\textbf{20.53} \\
 			\midrule
 			Time (s)  & 13.5 &  4.7  	 &	2.3		 & 21.5	      	  &	1.8	& 0.8 &		4.4 & 0.1 & 0.2 
 		\end{tabular}
 		\label{tab:full_pipeline_sup}
 		\egroup
 	\end{table*}
 }
 
\subsection{Results on metallic parts}
\begin{figure}[t]
	\centering
	\captionsetup{width=1.0\columnwidth}
	\includegraphics[width=1.0\columnwidth]{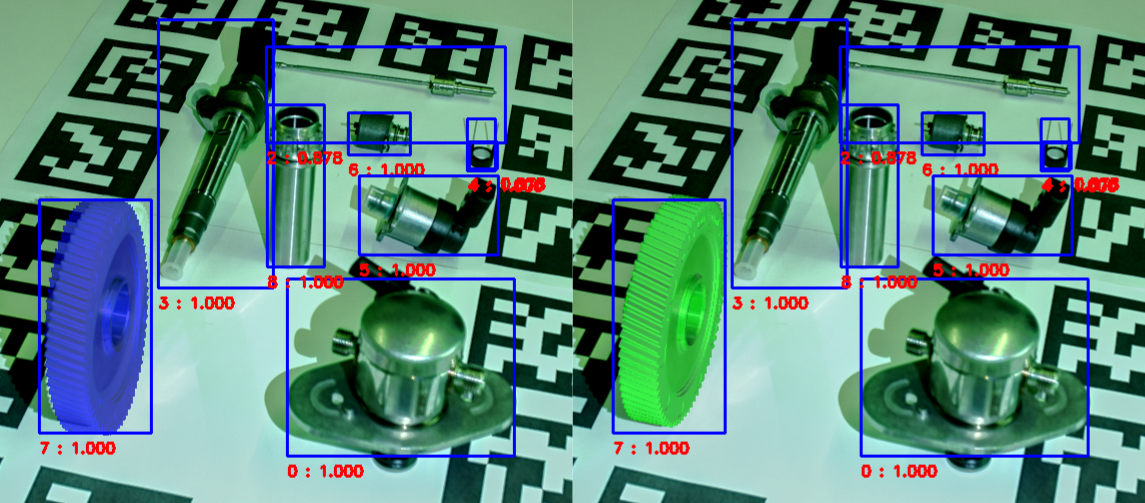}
	\caption{Left: MP-Encoder trained on T-LESS objects 1-18, and tested on a metallic, industrial object; Right: ICP-refined result}
	\label{fig:bosch}
\end{figure}

We briefly tested the MP-Encoder trained on T-LESS objects on metallic objects from industry (Fig. \ref{fig:bosch}).

 \subsection{Full T-LESS results}
 
\begin{figure*}[t]
	\centering
	\captionsetup{width=0.9\linewidth}
	\includegraphics[width=0.9\linewidth]{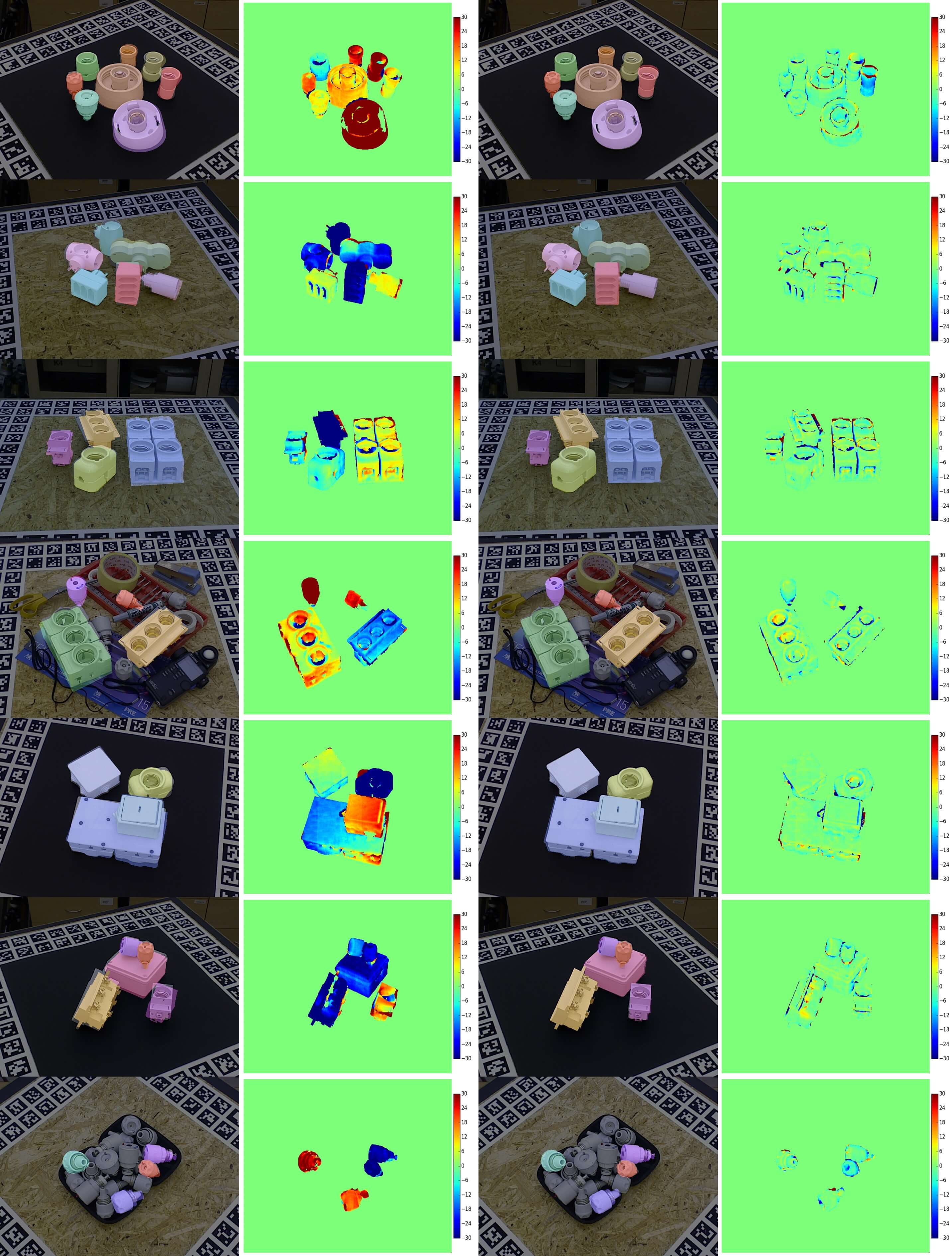}
	\caption{Qualitative 6D Object Detection results on T-LESS Primesense scenes. Only 18 of 30 objects are used to train the Multi-Path encoder. Left: Predictions of the RGB-based pipeline; Middle-left: Error in depth [mm]; Middle-right: ICP-refined results; Right: Refined error in depth [mm]. Note that only one instance per class is predicted following the BOP \cite{hodan2018bop} rules.}
	\label{fig:test_views}
\end{figure*}

 Table \ref{tab:full_pipeline} shows the full T-LESS results on each object tested on all scene views of the Primesense test set. Figure \ref{fig:test_views} show qualitative results with and without refinement on different T-LESS test scenes. Results on objects sometimes differ because the AAE submission uses RetinaNet (0.73mAP@0.5) and the MP-Encoder uses MaskRCNN (0.68mAP@0.5) because it does not learn to distinguish object and background (see Table 2).

\subsection{Runtime}

The MaskRCNN with ResNet50 takes $\sim150ms$ for all instances in a scene on a modern GPU. Lighter architectures can be chosen for simpler problems. All resulting image crops can be batched together such that the inference of the MP-Encoder is parallelized. For a single instance the inference takes $\sim5ms$, nearest neighbor search in the codebook takes $1-2ms$ and the projective distance estimation is negligible. While our RGB-based pipeline runs at high speed and makes our approach applicable for mobile applications, the depth-based \gls{ICP} refinement takes in average $0.6s$ per target and is thus suitable for robotic manipulation tasks.

%
%

\end{document}